\documentclass{article}
\pdfoutput=1

\usepackage[preprint,nonatbib]{neurips_2024}

\usepackage[utf8]{inputenc} 
\usepackage[T1]{fontenc}    
\usepackage{hyperref}       
\usepackage{url}            
\usepackage{booktabs}       
\usepackage{amsfonts}       
\usepackage{nicefrac}       
\usepackage{microtype}      
\usepackage{xcolor}         
\usepackage{graphicx}
\usepackage{wrapfig}
\usepackage{amsmath}

\usepackage{diagbox}
\usepackage{amsfonts,amssymb}
\usepackage{bbm}
\usepackage{multirow} 
\usepackage{amsthm}

\title{Continually Evolved Multimodal Foundation Models for Cancer Prognosis}

\author{%
  Jie Peng$^{1}$\thanks{Equal contribution.},\ Shuang Zhou$^{2}$\footnotemark[1],\ Longwei Yang$^{3}$,\ Yiran Song$^{2}$,\ Mohan Zhang$^{1}$, \\ \textbf{Kaixiong Zhou}$^{3}$,\ \textbf{Feng Xie}$^{2}$,\ \textbf{Mingquan Lin}$^{2}$,\ \textbf{Rui Zhang}$^{2\dagger}$,\ \textbf{Tianlong Chen}$^{1\dagger}$\\
  $^{1}$UNC at Chapel Hill, $^{2}$University of Minnesota, $^{3}$North Carolina State University\\
  $^\dagger$Corresponding to: \texttt{ruizhang@umn.edu} and \texttt{tianlong@cs.unc.edu} \\
}

\begin{document}

\maketitle

\begin{abstract}
Cancer prognosis is a critical task that involves predicting patient outcomes and survival rates. To enhance prediction accuracy, previous studies have integrated diverse data modalities, such as clinical notes, medical images, and genomic data, leveraging their complementary information. However, existing approaches face two major limitations. First, they struggle to incorporate newly arrived data with varying distributions into training, such as patient records from different hospitals, thus rendering sub-optimal generalizability and limited utility in real-world applications. Second, most multimodal integration methods rely on simplistic concatenation or task-specific pipelines, which fail to capture the complex interdependencies across modalities.
To address these, we propose a continually evolving multi-modal foundation model. Extensive experiments on the TCGA dataset demonstrate the effectiveness of our approach, highlighting its potential to advance cancer prognosis by enabling robust and adaptive multimodal integration.

\end{abstract}

\section{Introduction}

Cancer prognosis is a critical task that involves predicting the likely course and outcome of cancer, encompassing recovery probabilities, disease progression, and survival rates~\cite{zhu2020application}. As one of the leading causes of death worldwide, cancer poses a substantial barrier to extending life expectancy. For instance, the American Cancer Society reports nearly $0.7$ million cancer-related deaths annually~\cite{sung2021global}. Accurate prognostic assessments are essential for identifying high-risk patients who may benefit from aggressive interventions while sparing low-risk patients from unnecessary treatments and their associated side effects~\cite{zhu2020application}. Furthermore, robust prognostic tools play a pivotal role in clinical trial design, enabling better patient stratification and more effective evaluation of novel therapies.

Traditionally, cancer prognosis has been conducted by clinicians integrating diverse data sources, such as clinical notes, medical images, and genomic data~\cite{kawabata2015trends}. While these modalities collectively provide a comprehensive view of a patient’s condition, their sheer volume and heterogeneity present significant challenges. The high dimensionality and complexity of the data make manual analysis both time-intensive and prone to error. Additionally, the accuracy of prognostic assessments heavily depends on the clinician’s expertise, which can vary significantly across individuals and institutions. To address these limitations, computer-based methods leveraging artificial intelligence (AI) techniques have emerged as powerful tools to enhance the efficiency and accuracy of cancer prognosis~\cite{du2023multimodal, zhang2024mbfusion}.

Early approaches often relied on single data modalities for cancer prognosis~\cite{kourou2015machine}. However, subsequent studies demonstrated that integrating multiple data modalities~\cite{du2023multimodal, ding2023pathology}, such as clinical notes, medical images, and genomic data, yields superior performance by harnessing the complementary information these sources provide. Recently, the advent of multimodal foundation models has further advanced cancer prognosis by leveraging the paradigm shift in AI towards large-scale, unified models~\cite{wang2024pathology, kim2024llm}. These models integrate diverse data modalities within a cohesive framework, enabling more comprehensive analyses and accurate predictions. For example, TITAN~\cite{ding2024multimodal} is a multimodal foundation model pre-trained on an extensive dataset comprising 335,645 whole-slide images (WSIs) paired with pathology reports and 423,122 captions.

Despite these advancements, current efforts face two key limitations. 
First, real-world clinical environments frequently introduce new modalities or tasks with varying distributions, such as new patients from different hospitals, to boost model generalizability and utility~\cite{lee2020clinical}. Adapting multi-modality models to effectively incorporate these novel data sources for cancer prognosis remains an unresolved challenge. 
Second, existing multimodal integration approaches often rely on simplistic concatenation or task-specific pipelines~\cite{rahman2020integrating, tsai2019multimodal}, which fail to fully capture the intricate interdependencies across modalities. 
Addressing these limitations is crucial for further improving the applicability and robustness of cancer prognosis systems in diverse clinical settings.

In this study, we propose a continually evolving multi-modal foundation model to address these challenges in cancer prognosis. The model incorporates two key components: the Pseudo Target Generation Module (PTGM) and Instruction-based Knowledge Distillation (IKD). The PTGM module mitigates catastrophic forgetting when models are trained on multiple tasks within the same modality, while the IKD module preserves the model's generative capabilities when adapting to new modalities.
Extensive experiments conducted on the TCGA dataset~\cite{cancer2008comprehensive} validate the effectiveness of our proposed approach. The results underscore the potential of continually evolving multi-modal foundation models to significantly improve cancer prognosis in real-world clinical settings.

\section{Related Work}

\noindent\textbf{Multimodal Learning.} Multimodal Learning represents a critical frontier in artificial intelligence, focusing on integrating information across diverse modalities to enhance computational understanding and learning capabilities.
Recent advancements have been pioneering. Vaswani et al.~\cite{vaswani2017attention} introduced the Transformer architecture, providing a fundamental technological basis for multimodal learning, particularly through attention mechanisms enabling complex cross-modal associations. Subsequently, CLIP (Contrastive Language-Image Pre-training)~\cite{radford2021learning} revolutionized multimodal representation learning through contrastive learning techniques.
In the visual-language domain, models like ViLBERT~\cite{lu2019vilbert} and LXMERT~\cite{tan2019lxmert}proposed innovative multimodal fusion strategies, significantly improving performance in visual question-answering and image captioning tasks. More recent work, such as Flamingo~\cite{alayrac2022flamingo} and GPT-4V~\cite{openai2023gpt4v}, demonstrates the immense potential of large-scale multimodal models in handling complex cross-modal challenges.
Multimodal learning research extends beyond visual-linguistic domains~\cite{zhou2024large}, encompassing medical imaging~\cite{xue2022multimodal}, question answering~\cite{dong2024modality}, and HCI~\cite{kim2022advances}, indicating its transformative potential across technological landscapes.

\noindent\textbf{Foundation Model.} Foundation models (FMs) represent a transformative paradigm in artificial intelligence, characterized by large-scale pre-trained models capable of adapting to diverse downstream tasks~\cite{zhou2024large, bommasani2021foundation}. This approach has revolutionized machine learning across multiple domains, from natural language processing to computer vision and multimodal intelligence.
The conceptual groundwork was laid by seminal works in transfer learning and pre-training. The Transformer architecture~\cite{vaswani2017attention} fundamentally changed model design, introducing attention mechanisms that enabled more effective representation learning. Following this, models like BERT~\cite{devlin2018bert} and GPT series~\cite{radford2019gpt2,brown2020gpt3} demonstrated the power of large-scale pre-training on vast textual corpora.
Bommasani et al.~\cite{bommasani2021foundation} formally conceptualized "foundation models," highlighting their potential to serve as adaptable base models across multiple applications. The scaling laws proposed by Kaplan et al.~\cite{kaplan2020scaling} provided theoretical insights into model performance, demonstrating how model size and training data correlate with capabilities.
Multimodal foundation models have expanded the frontier of AI capabilities. CLIP~\cite{radford2021clip} revolutionized vision-language learning through contrastive pre-training, while models like DALL-E~\cite{ramesh2021dalle} and Stable Diffusion~\cite{rombach2022stable} enabled unprecedented text-to-image generation. GPT-4~\cite{openai2023gpt4} and GPT-4V~\cite{openai2023gpt4v} further pushed multimodal boundaries by integrating text, vision, and reasoning capabilities.
Specialized domain-specific foundation models have emerged in fields like biomedicine (AlphaFold~\cite{jumper2021alphafold}), scientific computing (PaLM~\cite{chowdhery2022palm}), and robotics (RT-1~\cite{brohan2022rt1}). These models demonstrate the potential of transfer learning in solving complex, specialized tasks.
Recent research has also focused on FMs' ethical implications and potential risks. Works by Bender et al.~\cite{bender2021dangers} and Weidinger et al.~\cite{weidinger2022ethical} critically examine these powerful AI systems' societal impacts, bias, and potential misuse. 
CREMA~\cite{yu2024crema} uses a shared Q-Former backbone across modalities to enhance parameter efficiency, employs a gated neural network for computational efficiency, and utilizes sequential training for training efficiency. 
Our method is inspired by CREMA but focuses on multi-modal fusion in medical applications. 
Additionally, we emphasize other properties of this fusion method, such as continual learning.

\noindent\textbf{Cancer Prognosis.} Early approaches to cancer prognosis primarily utilized machine learning techniques, such as support vector machines and Bayesian networks, or deep learning architectures, such as convolutional neural networks, typically relying on a single data modality~\cite{zhu2020application}. Subsequent studies demonstrated that fusing multiple data modalities~\cite{zhang2024mbfusion, du2023multimodal, ding2023pathology, liu2024pathformer, zhou2024multimodal}, including clinical notes, medical images, and genomic data, significantly enhanced performance by leveraging the complementary information inherent in different modalities. For example, Song et al.~\cite{songmultimodal} employed a Transformer to integrate gigapixel histology WSIs and transcriptomic profiles for survival prediction across six cancer types.
Recently, the paradigm shift in AI towards large-scale models has spurred the development of multimodal foundation models for cancer prognosis~\cite{wang2024pathology, kim2024llm, zhu2023samms}. These models unify diverse data modalities within a cohesive framework, enabling more comprehensive analyses and more accurate predictions~\cite{xu2024multimodal}. For instance, TITAN exemplifies this trend as a multimodal whole-slide foundation model, pre-trained on a dataset of 335,645 WSIs, the corresponding pathology reports, and 423,122 captions with supervised learning~\cite{ding2024multimodal}.
Despite these advancements, existing efforts face significant limitations in adapting to new modalities or tasks with varying distributions, such as data from different institutions or novel clinical scenarios.

\noindent\textbf{Continual Learning.} Continual learning (a.k.a., lifelong learning or incremental learning) is defined by its ability to learn from dynamic data distributions~\cite{de2021continual}. In practical scenarios, training samples with differing distributions arrive sequentially, posing unique challenges. A primary issue is catastrophic forgetting, where learning a new task often leads to significant performance degradation on previously learned tasks~\cite{shi2021overcoming}.
Traditional approaches to continual learning generally fall into five categories: regularization-based, replay-based, optimization-based, representation-based, and architecture-based methods~\cite{wang2024comprehensive}. More recently, the emergence of the foundation model era has spurred significant interest in enabling large models for continual learning~\cite{yang2024recent, shi2024continual}. The continual learning of foundation models typically involves multi-stage and cross-stage iterations~\cite{wu2024continual}, encompassing the following key aspects: (1) continual pre-training, which updates models with the latest information or adapts them to specialized domains~\cite{xie2023efficient}, (2) continual instruction-tuning, aimed at continually improving the ability to follow instructions and transfer knowledge to future tasks~\cite{mok2023large}, and (3) continual alignment, which ensures adaptation to evolving societal values, social norms, and ethical guidelines~\cite{zhang2023copf}.
A few pioneering studies~\cite{pian2024modality, zeng2024modalprompt, chen2024llm} have integrated continual learning into multimodal large language models (MLLMs), demonstrating their promising versatility in addressing diverse and adaptive tasks. Our work distinguishes itself from these efforts by being specifically tailored for cancer prognosis.
\section{Method}

\noindent\textbf{Multimodal Q-Former.} The Q-Former architecture~\cite{pmlr-v202-li23q} has demonstrated its ability to integrate multiple modalities with LLMs. However, these methods typically require a separate Q-Former for each modality, leading to an increase in parameter count that complicates scaling. To address this challenge, we introduce the \textbf{Multimodal Q-Former}, a lightweight and scalable architecture that integrates multiple modalities without the need for additional full-sized Q-Formers. The key innovation is the introduction of the \textit{Modality-specific Multi-Query Low-Rank Adaptation} (\textbf{MM-LoRA}) module which allowing each modality to be routed to its respective LoRA module while keeping the base Q-Former unchanged. Specifically, for each modality $m$, the MM-LoRA includes a set of learnable queries $\mathbf{q}^i_m$, linear projections, and each modality has its own modality-specific LoRA layers, which are uniquely tailored to that modality. The linear projection at each layer $i$ is computed as follows: 
\begin{equation}
    \mathbf{q}^{i+1}_m = \mathbf{W}_q \mathbf{q}^i_m + \Delta \mathbf{W}_m,
\end{equation}
where $\mathbf{W}_q$ indicates the original linear projection parameters of the Q-Former. The term $\Delta \mathbf{W}_m$ corresponds to the LoRA parameters specific to modality $m$, which is computed as:
\begin{equation}
\Delta \mathbf{W}_m = \mathbf{B}_m \mathbf{A}_m, \quad \mathbf{B}_m \in \mathbb{R}^{d \times r}, \, \mathbf{A}_m \in \mathbb{R}^{r \times d},
\end{equation}
where, $d$ is the embedding size of the Q-Former, and $r$ is the LoRA rank which is much smaller than $d$. When a new modality is introduced, we simply add a corresponding LoRA module, making it well-suited for multimodal continual learning.

\noindent\textbf{Self-gated Multimodal Query Fusion (SMQF).} To mitigate the computational overhead of LLM processing modality-specific query tokens from \textbf{MM-LoRA} modules, which are growing linearly with the number of modalities, we propose a \textbf{Self-Gated Multimodal Query Fusion (SMQF)} module. This module consolidates and compresses auxiliary modality queries into a unified representation while retaining the primary modality's dominance, thereby reducing computational costs and enabling the seamless addition of new modalities. Specifically, we first concatenate the supporting queries along the channel dimension, and project them into the same dimensional space as $\mathbf{x}_p$ using a linear transformation $\pi$:
\begin{equation}
\bar{\mathbf{x}}_{s} = \pi\left(\{\mathbf{x}_i\}_{i\neq p}^{s}; \mathbf{\theta}\right),
\end{equation}
where $n$ is the number of modalities, $p$ is the index of primary modality, and $\mathbf{x}_{s} = \{\mathbf{x}_i\}_{i=2}^n$ denotes the supporting queries of all modalities except the primary query $\mathbf{q}_p$. 
Next, we apply an element-wise sigmoid activation to $\bar{\mathbf{x}}_{s}$, followed by a Hadamard product to perform a \textbf{self-gating operation}, allowing the module to selectively focus on relevant features within the supporting modalities:
\begin{equation}
\mathbf{x}_{\textrm{gated}} = \textrm{sigmoid}(\bar{\mathbf{x}}_{s}) \odot \bar{\mathbf{x}}_{s},
\end{equation}
Finally, the gated supporting queries $\mathbf{x}_{\mathtt{gated}}$ are concatenated with the primary query $\mathbf{x}_p$ to form the \textbf{final multimodal query embedding}: $\mathbf{x} = \mathtt{concat}(\mathbf{x}_p, \mathbf{x}_{\mathtt{gated}})$.

\section{Experiment}

\subsection{Implement Details}

\noindent\textbf{Text preprocessing.}
We collected open-source pathology reports from the TCGA website~\cite{cancer2008comprehensive} and transformed them from their original PDF format into editable text using Amazon Web Services (AWS) Optical Character Recognition (OCR) tools. This process yielded 9,523 reports spanning 32 cancer types. To encode the raw text, we employed a text encoder with a feature dimension of 768, converting the text into token embeddings.

\noindent\textbf{Image data preprocessing.}
To ensure a focus on the methodology, we utilize identical feature extractors across all datasets, with all methods operating exclusively on vectorized data obtained after feature extraction. Specifically, we adopt a ViT architecture with pre-trained weights from Marugoto~\cite{marugoto}. Each WSI is divided into non-overlapping 224 × 224 patches to form a bag, while background patches (entropy < 5) are discarded. To standardize the number of patches extracted from WSI images, a predefined threshold is applied. Subsequently, all models are trained on the complete set of patches to ensure a fair comparison. For the survival analysis task, the input feature dimension is set to 2048.

\noindent\textbf{Genome preprocessing.}
We use the bulk RNA data collected from the TCGA dataset and the BulkRNABert with pre-trained weight from the original paper~\cite{gelard2024bulkrnabert}. Specifically, an RNA sequence is represented as a vector, where each dimension corresponds to the expression level of a specific gene. The vector is sent into the BulkRNBert model to generate an N × 256 tensor, and the embedding of the whole sequence will be the average of those N vectors. 

\subsection{Performance of Multi-modal Integration}

\begin{table}[!ht]
\caption{The multi-modal integration performance comparison.}
\label{tab:multi_modal_integration}
\resizebox{\textwidth}{!}{%
\begin{tabular}{c|c|c|c|c|c|c|c}
\toprule
\multicolumn{1}{l|}{} & Modality & UCEC & LUAD & LGG & BRCA & BLCA & MEAN \\ \midrule
LiMOE & Text, RNA, Image & 0.795 & \textbf{0.674} & 0.771 & 0.736 & 0.578 & 0.711 \\ \midrule
MAGGate & Text, RNA, Image & 0.613 & 0.621 & 0.799 & 0.684 & 0.475 & 0.638 \\ \midrule
MulT & Text, RNA, Image & 0.665 & 0.616 & 0.828 & 0.725 & 0.569 & 0.681 \\ \midrule
TF & Text, RNA, Image & 0.722 & 0.664 & 0.696 & 0.721 & 0.565 & 0.674 \\ \midrule
Cross Attn Fusion & Text, RNA, Image & {0.865} & 0.617 & \textbf{0.831} & 0.731 & 0.541 & 0.717 \\ \midrule
Early Fusion & Text, RNA, Image & 0.697 & 0.649 & 0.83 & 0.665 & 0.584 & 0.685 \\ \midrule
Late Fusion & Text, RNA, Image & 0.679 & 0.584 & 0.747 & 0.761 & 0.588 & 0.672 \\ \midrule
Cross Attn Fusion & Text, RNA, Image & 0.811 & 0.64 & 0.761 & 0.751 & 0.506 & 0.694 \\ \midrule
Early Fusion & Text, Image & \textbf{0.881} & 0.646 & 0.829 & 0.756 & 0.618 & 0.746 \\ \midrule
Late Fusion & Text, Image & 0.846 & 0.634 & 0.818 & 0.745 & 0.566 & 0.722 \\ \midrule
Text Only (BERT) & Text & 0.57 & 0.446 & 0.56 & 0.747 & 0.546 & 0.574 \\ \midrule
Text Only (MUSK) & Text & 0.785 & 0.649 & 0.751 & 0.719 & 0.645 & 0.710 \\ \midrule
RNA Seq Only (CustOmics) & RNA & 0.660 & 0.638 & 0.609 & 0.611 & 0.626 & 0.629 \\ \midrule
RNA Seq Only (BulkRNABert) & RNA & 0.703 & 0.621 & 0.775 & 0.604 & 0.627 & 0.666 \\ \midrule
RNA Seq Only & RNA & 0.787 & 0.431 & 0.765 & 0.651 & 0.639 & 0.655 \\ \midrule
Image Only (ABMIL) & Image & 0.651 & 0.525 & 0.493 & 0.502 & 0.47 & 0.528 \\ \midrule
Image Only (DSMIL) & Image & 0.754 & 0.541 & 0.681 & 0.633 & 0.572 & 0.636 \\ \midrule
Image Only (ESAT) & Image & 0.723 & 0.649 & 0.638 & 0.544 & 0.545 & 0.620 \\ \midrule
CREMA & Text, RNA, Image & 0.863 & 0.658 & 0.687 & \textbf{0.781} & \textbf{0.775} & \textbf{0.753} \\ \bottomrule
\end{tabular}%
}
\end{table}

We evaluate our proposed multi-modal fusion mechanism against existing methods, such as early fusion, late fusion, cross-attention fusion, and tensor fusion. We also compare our approach with recent advanced methods like MulT~\cite{tsai2019multimodal} and MAGGate~\cite{rahman2020integrating}. MulT uses modality-specific cross-attention to integrate other modalities with the current one, then concatenates all fused features for final prediction. MAGGate uses a gated network to merge all modalities. Our evaluation focuses on the cancer prognosis task across various cancer types, including UCEC, LUAD, LGG, BRCA, and BLCA from the TCGA dataset.

As shown in Table~\ref{tab:multi_modal_integration}, our method achieves average c-index improvements of \{0.042, 0.115, 0.072, 0.079, 0.036, 0.068, 0.081\} over \{LiMOE, MAGGate, MulT, TF, Cross Attn Fusion, Early Fusion, Late Fusion\} baselines. This indicates that our method consistently enhances performance across different cancer types. Compared to recent state-of-the-art single-modal baselines, we also achieve the best performance on \{UCEC, LUAD, BRCA, BLCA\}, with c-index improvements of \{0.07$\sim$0.293, 0.009$\sim$0.227, 0.034$\sim$0.279, 0.136$\sim$0.305\}. Additionally, we observe average performance improvements of \{0.179, 0.098, 0.117$\sim$0.225\} compared to the \{Text Only, RNA-seq Only, Image Only\} baselines.

\subsection{Performance of Continual Learning}


\begin{wrapfigure}{r}{0.5\linewidth}
    \centering
    \includegraphics[width=0.98\linewidth]{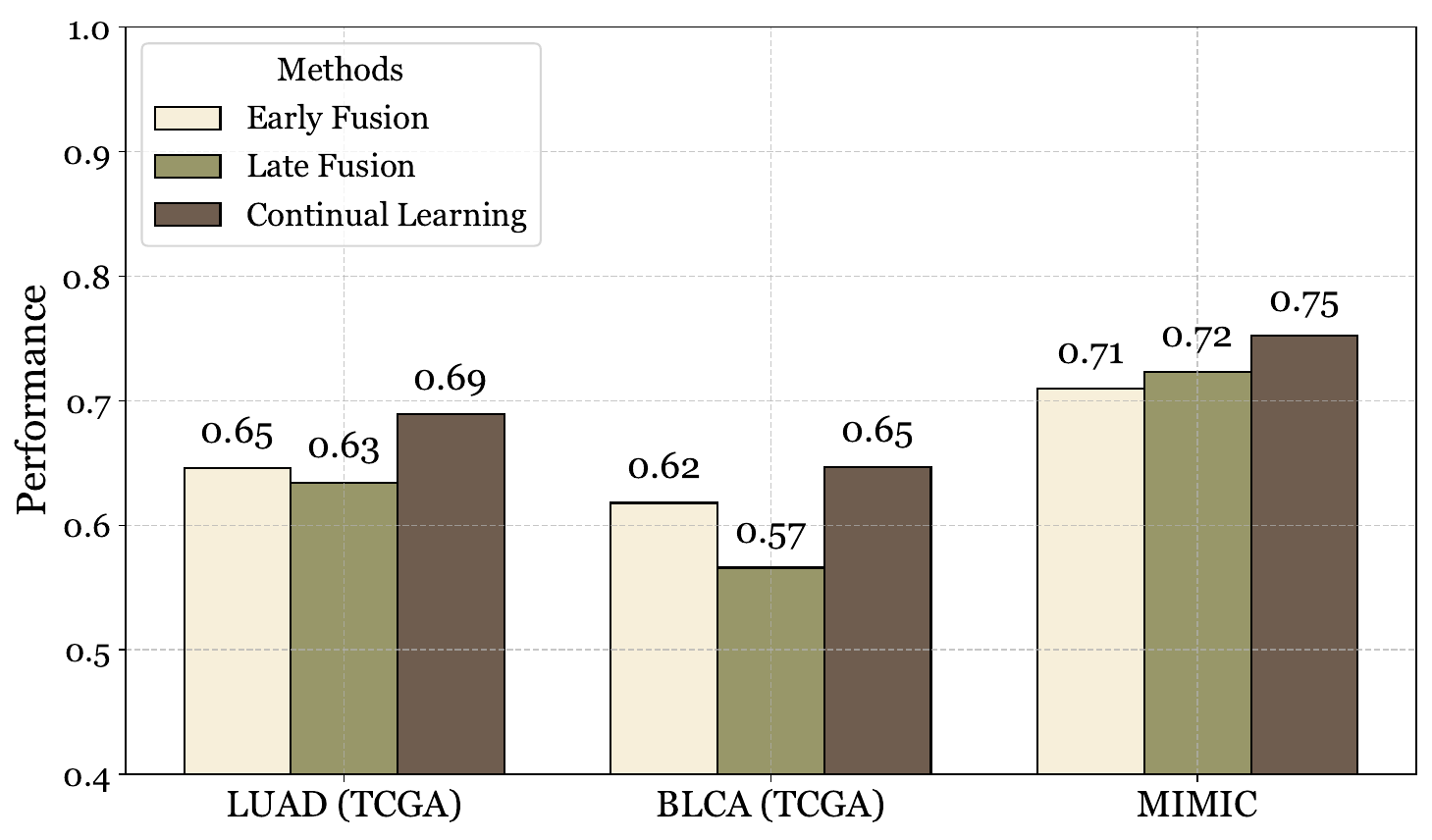}
    \vspace{-4mm}
    \caption{Performance of Continual Learning.}
    \label{fig:prefilling}
    \vspace{-4mm}
\end{wrapfigure}

We validated our continual learning design using both TCGA and MIMIC datasets. Specifically, we focused on two cancer types from the TCGA dataset: LUAD and BLCA. The model was initially trained with image and clinical text modalities, then continued learning with genomics modalities. For the MIMIC dataset, the model started with code and clinical notes modalities and then incorporated the lab modality. We employed two classic multi-modal learning methods: early fusion and late fusion. As shown in Figure~\ref{fig:prefilling}, our method effectively learns new modalities and improves testing c-index performance by up to \{$0.055$ for LUAD, $0.081$ for BLCA, $0.042$ for MIMIC\}.

\section{Conclusion}

In this study, we propose a continually evolved multi-modal foundation model to incorporate new data into training and effectively capture intricate interdependence across modalities for cancer prognosis. 
Experiments on the TCGA dataset verified the effectiveness of our method. Future studies can further explore more efficient and scalable algorithms for large-scale data or validate the cancer prognosis performance in real-world scenarios.

\addcontentsline{toc}{section}{References}
\bibliography{main}   
\bibliographystyle{unsrt}











\end{document}